\documentclass[10pt,twocolumn,letterpaper]{article}

\usepackage{cvpr}
\usepackage{times}
\usepackage{epsfig}
\usepackage{graphicx}
\usepackage{amsmath}
\usepackage{amssymb,shortbold}
\usepackage{verbatim,booktabs}
\usepackage{xcolor,colortbl}

\usepackage[pagebackref=true,breaklinks=true,letterpaper=true,colorlinks,bookmarks=false]{hyperref}

\cvprfinalcopy 

\def\cvprPaperID{1616} 

\definecolor{CTX0b0}{rgb}{0.993248,0.906157,0.143936}
\definecolor{CTX0f0}{rgb}{0.000000,0.000000,0.000000}
\definecolor{CTX0b1}{rgb}{0.993248,0.906157,0.143936}
\definecolor{CTX0f1}{rgb}{0.000000,0.000000,0.000000}
\definecolor{CTX0b2}{rgb}{0.876168,0.891125,0.095250}
\definecolor{CTX0f2}{rgb}{0.000000,0.000000,0.000000}
\definecolor{CTX0b3}{rgb}{0.993248,0.906157,0.143936}
\definecolor{CTX0f3}{rgb}{0.000000,0.000000,0.000000}
\definecolor{CTX0b4}{rgb}{0.814576,0.883393,0.110347}
\definecolor{CTX0f4}{rgb}{0.000000,0.000000,0.000000}
\definecolor{CTX0b5}{rgb}{0.845561,0.887322,0.099702}
\definecolor{CTX0f5}{rgb}{0.000000,0.000000,0.000000}
\definecolor{CTX0b6}{rgb}{0.993248,0.906157,0.143936}
\definecolor{CTX0f6}{rgb}{0.000000,0.000000,0.000000}
\definecolor{CTX1b0}{rgb}{0.668054,0.861999,0.196293}
\definecolor{CTX1f0}{rgb}{0.000000,0.000000,0.000000}
\definecolor{CTX1b1}{rgb}{0.730889,0.871916,0.156029}
\definecolor{CTX1f1}{rgb}{0.000000,0.000000,0.000000}
\definecolor{CTX1b2}{rgb}{0.585678,0.846661,0.249897}
\definecolor{CTX1f2}{rgb}{0.000000,0.000000,0.000000}
\definecolor{CTX1b3}{rgb}{0.783315,0.879285,0.125405}
\definecolor{CTX1f3}{rgb}{0.000000,0.000000,0.000000}
\definecolor{CTX1b4}{rgb}{0.506271,0.828786,0.300362}
\definecolor{CTX1f4}{rgb}{0.000000,0.000000,0.000000}
\definecolor{CTX1b5}{rgb}{0.606045,0.850733,0.236712}
\definecolor{CTX1f5}{rgb}{0.000000,0.000000,0.000000}
\definecolor{CTX1b6}{rgb}{0.668054,0.861999,0.196293}
\definecolor{CTX1f6}{rgb}{0.000000,0.000000,0.000000}
\definecolor{CTX2b0}{rgb}{0.223925,0.334994,0.548053}
\definecolor{CTX2f0}{rgb}{0.900000,0.900000,0.900000}
\definecolor{CTX2b1}{rgb}{0.195860,0.395433,0.555276}
\definecolor{CTX2f1}{rgb}{0.900000,0.900000,0.900000}
\definecolor{CTX2b2}{rgb}{0.993248,0.906157,0.143936}
\definecolor{CTX2f2}{rgb}{0.000000,0.000000,0.000000}
\definecolor{CTX2b3}{rgb}{0.157851,0.683765,0.501686}
\definecolor{CTX2f3}{rgb}{0.900000,0.900000,0.900000}
\definecolor{CTX2b4}{rgb}{0.220057,0.343307,0.549413}
\definecolor{CTX2f4}{rgb}{0.900000,0.900000,0.900000}
\definecolor{CTX2b5}{rgb}{0.274128,0.199721,0.498911}
\definecolor{CTX2f5}{rgb}{0.900000,0.900000,0.900000}
\definecolor{CTX2b6}{rgb}{0.271828,0.209303,0.504434}
\definecolor{CTX2f6}{rgb}{0.900000,0.900000,0.900000}
\definecolor{CTX3b0}{rgb}{0.144759,0.519093,0.556572}
\definecolor{CTX3f0}{rgb}{0.900000,0.900000,0.900000}
\definecolor{CTX3b1}{rgb}{0.172719,0.448791,0.557885}
\definecolor{CTX3f1}{rgb}{0.900000,0.900000,0.900000}
\definecolor{CTX3b2}{rgb}{0.129933,0.559582,0.551864}
\definecolor{CTX3f2}{rgb}{0.900000,0.900000,0.900000}
\definecolor{CTX3b3}{rgb}{0.993248,0.906157,0.143936}
\definecolor{CTX3f3}{rgb}{0.000000,0.000000,0.000000}
\definecolor{CTX3b4}{rgb}{0.244972,0.287675,0.537260}
\definecolor{CTX3f4}{rgb}{0.900000,0.900000,0.900000}
\definecolor{CTX3b5}{rgb}{0.280868,0.160771,0.472899}
\definecolor{CTX3f5}{rgb}{0.900000,0.900000,0.900000}
\definecolor{CTX3b6}{rgb}{0.279574,0.170599,0.479997}
\definecolor{CTX3f6}{rgb}{0.900000,0.900000,0.900000}
\definecolor{CTX4b0}{rgb}{0.232815,0.732247,0.459277}
\definecolor{CTX4f0}{rgb}{0.900000,0.900000,0.900000}
\definecolor{CTX4b1}{rgb}{0.288921,0.758394,0.428426}
\definecolor{CTX4f1}{rgb}{0.900000,0.900000,0.900000}
\definecolor{CTX4b2}{rgb}{0.657642,0.860219,0.203082}
\definecolor{CTX4f2}{rgb}{0.000000,0.000000,0.000000}
\definecolor{CTX4b3}{rgb}{0.252899,0.742211,0.448284}
\definecolor{CTX4f3}{rgb}{0.900000,0.900000,0.900000}
\definecolor{CTX4b4}{rgb}{0.993248,0.906157,0.143936}
\definecolor{CTX4f4}{rgb}{0.000000,0.000000,0.000000}
\definecolor{CTX4b5}{rgb}{0.886271,0.892374,0.095374}
\definecolor{CTX4f5}{rgb}{0.000000,0.000000,0.000000}
\definecolor{CTX4b6}{rgb}{0.327796,0.773980,0.406640}
\definecolor{CTX4f6}{rgb}{0.000000,0.000000,0.000000}
\definecolor{CTX5b0}{rgb}{0.153894,0.680203,0.504172}
\definecolor{CTX5f0}{rgb}{0.900000,0.900000,0.900000}
\definecolor{CTX5b1}{rgb}{0.288921,0.758394,0.428426}
\definecolor{CTX5f1}{rgb}{0.900000,0.900000,0.900000}
\definecolor{CTX5b2}{rgb}{0.545524,0.838039,0.275626}
\definecolor{CTX5f2}{rgb}{0.000000,0.000000,0.000000}
\definecolor{CTX5b3}{rgb}{0.239374,0.735588,0.455688}
\definecolor{CTX5f3}{rgb}{0.900000,0.900000,0.900000}
\definecolor{CTX5b4}{rgb}{0.935904,0.898570,0.108131}
\definecolor{CTX5f4}{rgb}{0.000000,0.000000,0.000000}
\definecolor{CTX5b5}{rgb}{0.993248,0.906157,0.143936}
\definecolor{CTX5f5}{rgb}{0.000000,0.000000,0.000000}
\definecolor{CTX5b6}{rgb}{0.214000,0.722114,0.469588}
\definecolor{CTX5f6}{rgb}{0.900000,0.900000,0.900000}

\definecolor{SC0b0}{rgb}{0.993248,0.906157,0.143936}
\definecolor{SC0f0}{rgb}{0.000000,0.000000,0.000000}
\definecolor{SC0b1}{rgb}{0.993248,0.906157,0.143936}
\definecolor{SC0f1}{rgb}{0.000000,0.000000,0.000000}
\definecolor{SC0b2}{rgb}{0.993248,0.906157,0.143936}
\definecolor{SC0f2}{rgb}{0.000000,0.000000,0.000000}
\definecolor{SC0b3}{rgb}{0.993248,0.906157,0.143936}
\definecolor{SC0f3}{rgb}{0.000000,0.000000,0.000000}
\definecolor{SC0b4}{rgb}{0.993248,0.906157,0.143936}
\definecolor{SC0f4}{rgb}{0.000000,0.000000,0.000000}
\definecolor{SC0b5}{rgb}{0.993248,0.906157,0.143936}
\definecolor{SC0f5}{rgb}{0.000000,0.000000,0.000000}
\definecolor{SC1b0}{rgb}{0.964894,0.902323,0.123941}
\definecolor{SC1f0}{rgb}{0.000000,0.000000,0.000000}
\definecolor{SC1b1}{rgb}{0.886271,0.892374,0.095374}
\definecolor{SC1f1}{rgb}{0.000000,0.000000,0.000000}
\definecolor{SC1b2}{rgb}{0.945636,0.899815,0.112838}
\definecolor{SC1f2}{rgb}{0.000000,0.000000,0.000000}
\definecolor{SC1b3}{rgb}{0.945636,0.899815,0.112838}
\definecolor{SC1f3}{rgb}{0.000000,0.000000,0.000000}
\definecolor{SC1b4}{rgb}{0.575563,0.844566,0.256415}
\definecolor{SC1f4}{rgb}{0.000000,0.000000,0.000000}
\definecolor{SC1b5}{rgb}{0.506271,0.828786,0.300362}
\definecolor{SC1f5}{rgb}{0.000000,0.000000,0.000000}
\definecolor{SC2b0}{rgb}{0.751884,0.874951,0.143228}
\definecolor{SC2f0}{rgb}{0.000000,0.000000,0.000000}
\definecolor{SC2b1}{rgb}{0.595839,0.848717,0.243329}
\definecolor{SC2f1}{rgb}{0.000000,0.000000,0.000000}
\definecolor{SC2b2}{rgb}{0.595839,0.848717,0.243329}
\definecolor{SC2f2}{rgb}{0.000000,0.000000,0.000000}
\definecolor{SC2b3}{rgb}{0.565498,0.842430,0.262877}
\definecolor{SC2f3}{rgb}{0.000000,0.000000,0.000000}
\definecolor{SC2b4}{rgb}{0.124780,0.640461,0.527068}
\definecolor{SC2f4}{rgb}{0.900000,0.900000,0.900000}
\definecolor{SC2b5}{rgb}{0.125394,0.574318,0.549086}
\definecolor{SC2f5}{rgb}{0.900000,0.900000,0.900000}

\ifcvprfinal\fi
\begin{document}

\title{Understanding Human Hands in Contact at Internet Scale}

\author{Dandan Shan$^1$, Jiaqi Geng*$^1$, Michelle Shu*$^2$, David F. Fouhey$^1$\\
$^{1}$University of Michigan, $^{2}$Johns Hopkins University\\
{\tt \small \{dandans,jiaqig,fouhey\}@umich.edu, msh1@jhu.edu}
}
\newcommand{\review}[1]{{\huge \bf{\textcolor{orange}{#1}}}}

\newcommand{\note}[1]{}
\newcommand{\dandan}[1]{{}}

\maketitle

\begin{abstract}
Hands are the central means by which humans manipulate their
world and being able to reliably extract hand state information from
Internet videos of humans engaged in their hands has the potential
to pave the way to systems that can learn from petabytes of video data. 

This paper proposes steps towards this by inferring a rich representation of hands
engaged in interaction method that includes: hand
location, side, contact state, and a box around the object in contact.
To support this effort, we gather a large-scale dataset of hands in contact with
objects consisting of 131 days of footage as well as a 100K annotated
hand-contact video frame dataset. The learned model on this dataset can
serve as a foundation for hand-contact understanding in videos. 
We quantitatively evaluate it both on its own and in service of predicting
and learning from 3D meshes of human hands. 
\end{abstract}

\section{Introduction}
\label{sec:introduction}
\begin{figure}
\centering
\includegraphics[width=\linewidth]{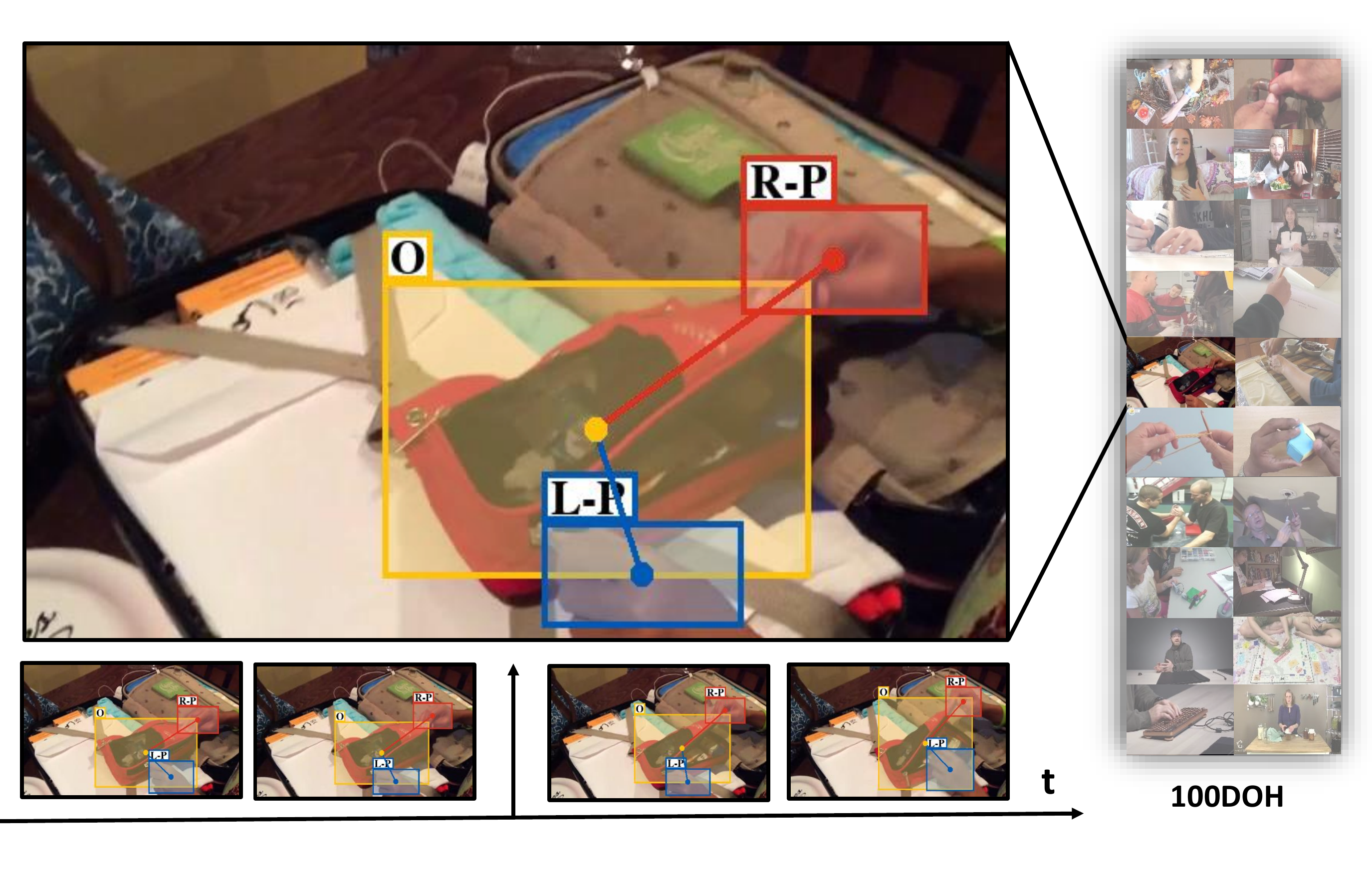}
\caption{The goal of this paper is to infer a rich representation for
helping understand hands engaged in contact with the world at Internet scale.
Our system produces a rich output in terms of hand location (boxes), side (left/right),
contact state (here -- a portable object) and what object each hand is in contact with.
To support this, we collect a new large-scale hand video interaction dataset, 100 Days of Hands 
({\it 100DOH}). We use this, plus videos from VLOG \cite{Fouhey18}, to make a 100K image dataset 
annotated with our rich representation. \dandan{R1: difficult to see. Will do.}
\note{df: can ignore this}
\label{fig:teaser}
}
\end{figure}

The hand is the key to how humans interact with the world.
If machines are to understand our actions and intentions as well as the
world we have build for and with our hands, they must have a deep understanding of our hands.
For instance, in Figure \ref{fig:teaser}, we can readily recognize that there are two hands
(one left and one right), opening a bag and even imagine how one might pull up the
flap. The goal of this paper is to build the foundation for studying hands engaged
in interaction with objects at Internet scale.

Hand analysis is, of course, an area of long-standing interest in the field
with work on pose estimation \cite{Zimmerman2017,Spurr2018}, reconstruction
\cite{Hasson19,Freihand2019}, and grasp analysis \cite{Rogez15,Brahmbhatt19}.
These approaches, however, have largely focused on in-lab settings, often with
a pre-localized hand or in settings with limited variety. While there has been substantial
progress, deploying these on the rich world of Internet videos
\cite{Alayrac2016,Zhou2018} poses a challenge due to the dizzying
diversity in viewpoint and context. A single system must handle
data ranging from a fifty pixel high hand in a cooking
video to an enormous thousand pixel high hand closeup showing DIY.


\begin{figure*}[!t]
\centering
\includegraphics[width=\linewidth]{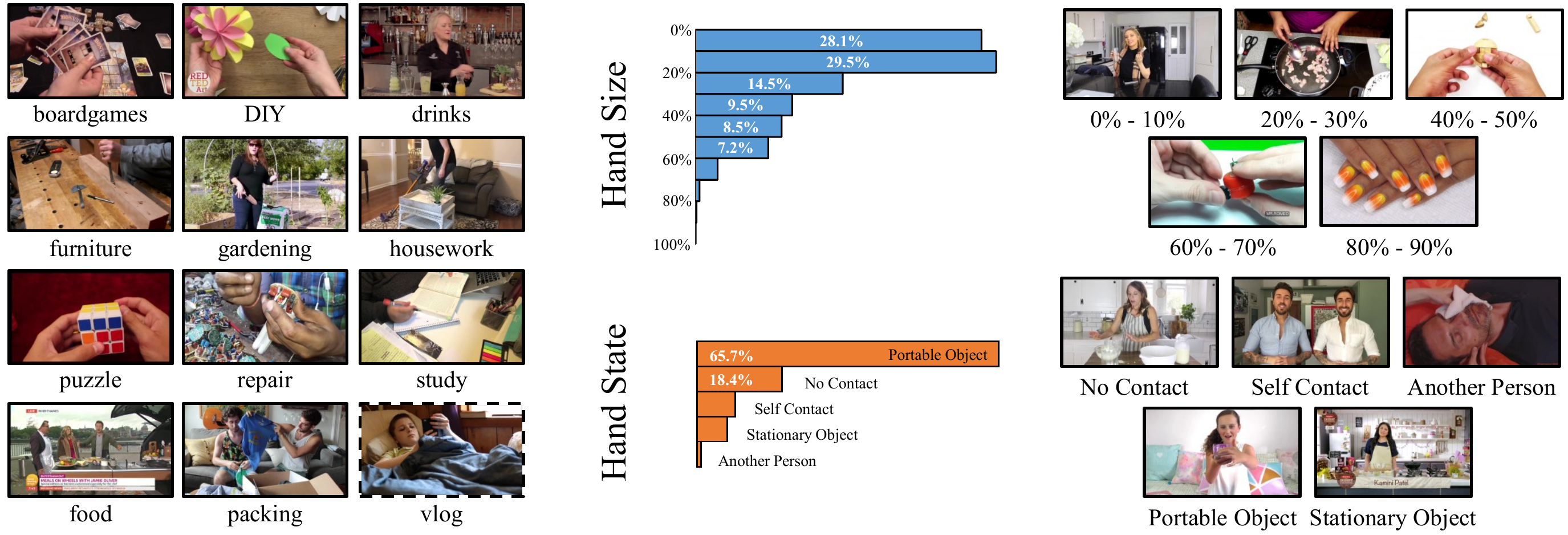}
    \caption{{\bf Snapshots from our dataset.} {\bf (Left)} Samples from the eleven genres that we use
    to collect our video dataset {\it 100DOH} as well as a sample from the VLOG dataset that we additionally
    sample from to produce our 100K frame subset. {\bf (Right)} Statistics
    about the hand frame dataset in terms of hand size (bounding box diagonal
    length over image diagonal length) and contact state, as well as
    illustrative samples of both. Our dataset depicts a wide
    variety of hands engaged in interaction in a variety of contexts
    and at a variety of scales.}
\end{figure*}

Our work aims to enable hand analysis at Internet scale and diversity. 
To this end, we introduce a model, described in Section
\ref{sec:method} that identifies, for every single hand in a single RGB image
(demonstrated on a wide variety of scales and contexts): a {\it hand box}; its {\it side}
(left/right); its {\it contact state} (none / self / other person /
non-portable object / portable object); and, for the hand in contact, an {\it object box} around
the object or person in contact. 
These enable crucial downstream problems
like reconstruction and grasp analysis. For example, the detection of
hand location and side enables the use of recent mesh reconstruction systems
\cite{Hasson19,Freihand2019}.

This system, along with an existing mesh reconstruction method \cite{Hasson19},
yields a system that can detect hands, their contact state, their 3D
reconstruction, and what object they are touching. We believe this output
enables large-scale fine-grained learning about human-object
interaction. As an example, we introduce a method to identify bad mesh
reconstructions (necessary for learning {\it from} meshes) and provide
a demonstration of learning on consumer videos.


This effort is backed by a new dataset, {\it 100 Days of
Hands}, introduced in Section \ref{sec:dataset}, consisting of a large-scale
(131 days+) video dataset of humans engaged in interaction that was gathered
implicitly \cite{Fouhey18} (i.e., with a set of relatively generic genre tags
as opposed to particular actions). These videos depict a wide range of activities,
viewpoints, and settings. We use frames from this dataset and 
a similar dataset \cite{Fouhey18} to create a 100K image
dataset which has the rich hand-state annotation to support our model. 

We believe that this model and data (along with existing past work,
especially in reconstruction) enable the field to collectively tackle new and important
problems in human-object interaction in general consumer videos and 
demonstrate this concretely in Section \ref{sec:experiments}. We show that:
(1) Our new hand dataset yields high performance detectors (90\% VOC \cite{Everingham10} AP)
that generalize well across datasets, occasionally outperforming training and testing on the
same dataset;
(2) Our new hand state model and dataset serves as an enabling technology that lets
the community deploy exciting hand-mesh reconstruction systems like \cite{Hasson19} on 
YouTube videos;
(3) Our system may provide a stepping stone towards important tasks like grasp analysis
by showing how to use it to build a proof-of-concept system that maps objects to 3D meshes
of hands engaged in interaction.

\section{Related Work}
\label{sec:related}

Our work focuses on identifying a rich state of hands in an ordinary RGB image
with the goal of using it as a foundation for understanding human-object
interaction in Internet videos. It therefore touches on many papers in the area
of understanding human-object interaction.


Human-object interaction, and understanding the affordances (opportunities
for interaction \cite{Gibson79}) has been a long-term interest of computer
vision. Recent work has largely taken the approach of recognizing 
verb-noun pairs \cite{Gkioxari18,Chao2018,Chao15,Gupta15,Baradel18}. In terms
of technical approach, our method is most related to the approach of Gkioxari et al \cite{Gkioxari18}. 
As an output, however, we propose an alternate representation based on physical
contact and interaction. 

In the process, we gather a video dataset of humans engaged in interaction that we
annotate and learn from. Of the many works in video human-object interaction
\cite{Damen18,Zhou2018,Sigurdsson2016,miech2019howto100m,Zhukov19}, ours 
is most clearly related to VLOG \cite{Fouhey18}, which also gathers 
data of interaction, and AVA \cite{Gu2018}, which
investigates atomic (i.e., base-level) actions. We build on the ideas
and part of the data of VLOG, but expand it to a wider and more diverse
dataset and far more thoroughly investigate and annotate contact.
Like AVA, we also investigate a representation below activities, but is
different and complementary (in contact with a box) compared to AVA's semantic
actions (e.g., ``write'', ``play instrument'').

%

Hands have long been the subject of study in computer vision. In this area,
our data and approach sit between image-based hand detection datasets
like \cite{VIVA,Mittal11,Bambach15,Fouhey18,Narasimhaswamy2019} and efforts at understanding
contact with richer annotation but requiring more specialized devices or more constrained environments such as
\cite{Brahmbhatt19,Sridhar2016,Rogez15,GarciaHernando18,Tekin19}. We expect
that fully understanding hands may require a variety of approaches; our 
approach tries to strike a balance between potential for scalability and
richness of annotation.
We note that while there has been progress
in full body pose estimation (e.g., \cite{Cao2017}), our approach
works even in highly truncated settings like Internet videos.

One particularly important line of work in understanding human hands is 
extracting the pose of hands from images. A full survey of this 
literature is beyond the scope of this paper and we refer the reader to
\cite{Zimmerman2017,Spurr2018}. Most recently, this has taken the form of
systems that can, given a cropped hand with known side, such as
\cite{Hasson19,Freihand2019}, infer a mesh via a low-dimensional model like MANO
\cite{Romero17}.  Our approach provides the necessary input for this
reconstruction and thus enables the large-scale deployment of these techniques
to Internet videos (including a self-supervised learning-based system that can
detect reconstruction failures). 

One of the applications we demonstrate is mapping an image of an object
not being interacted with to a 3D mesh of the hand. This work has tackled
previously using RGBD sensors \cite{Hamer2010,Akizuki2018} or 
thermal data \cite{Brahmbhatt19}; our work is able to learn this simply by
mining examples via the rich representation our approach can infer. Most
existing work in this area that can learn from Internet videos
\cite{Nagarajan2019,Fang2018} focuses on interaction hotspots, while our work
infers a mesh.

\section{Dataset}
\label{sec:dataset}
\begin{table}[t]
    \centering
    \caption{Comparison of 100DOH with existing datasets for human-object interaction. 
    While only a small fraction of it is labeled compared to more densely annotated datasets, the proposed large-scale
    video dataset is a rich source for unsupervised learning about hands. \dandan{Table 1 is Unfair, need to change a way to compare, to distinguish raw video data and total annotations.}
    }
    \label{tab:datacompare}
    {\footnotesize
    \begin{tabular}{@{~}l@{~~~}c@{~~~}c@{~~~}c@{~}} \toprule
        Name & Length & Annotation & Source \\ \midrule
        100DOH & 131D & 100K frames per-hand state & YouTube
        \\
        AVA & 2D & 3s-level atomic Actions & Movies 
        \\
        HowTo100M \cite{miech2019howto100m} & (5.6K)D & None / Captions & YouTube
        \\
        Moments \cite{MonfortMoments2019} & 34D, 17H & Vid. Class & Misc 
        \\
        VLOG \cite{Fouhey18} &  14D, 8H & Vid Class, Sparse Annots & YouTube
        \\
        YouCook2 \cite{Zhou2018} & 7D, 8H & Action Segments & YouTube
        \\
        Charades \cite{Sigurdsson2016} & 3D, 8H & Action Segments & Home
        \\
        EPIC-KITCH. \cite{Damen18} & 2D, 7H & Actions Segments, Object BBs & Home
        \\
        \bottomrule
    \end{tabular}}
\end{table}

We gathered a large and rich dataset of everyday interactions
from YouTube that serves as a basis for our subsequent investigation.
This dataset consists of two parts that play complementary roles: (i) a
massive, unlabeled video dataset that is source for
unsupervised learning; and (ii) a 100K frame subset that has
been labeled.

We follow the principles outlined in \cite{Fouhey18},
where we search {\it implicitly} for hands engaged in interactions
rather than explicitly. 
We see a few advantages to frames from implicitly gathered
video data: (a) still photos require an intentional decision to
take and upload the photo, meaning that the transitional fossils of daily life
(e.g., a half-ajar refrigerator with a hand resting on it) usually go
undocumented, a form of selection bias \cite{Torralba2011}; (b) explicitly
gathered data (e.g., searching ``playing tuba'' for Kinetics
\cite{Kay17}) tends to capture unusual activities since these are easy to find.

\subsection{Gathering a Large-Scale Video Dataset}

Gathering implicitly consists of two rough stages: identifying an
overcomplete set of candidate videos using generic
queries and filtering out irrelevant videos.

\vspace{2mm}
\noindent {\bf Generating a set of query candidates:}
We began with a set of 11 categories: boardgames, DIY, making drinks, making
food, furniture assembly, gardening, doing housework, packing, doing puzzles,
repairing, and studying. We generated 13.2K queries using frequent words, Wordnet
hyponyms and templated queries (e.g., ``DIY cookies home 2014''), and searched
YouTube. These queries yield $\sim$6.5M video responses (an estimated 
86 years), which we must filter for containing hands interacting
with objects.

\vspace{2mm}
\noindent {\bf Filtering:} Manually screening such a large dataset is 
impractical and we therefore use a learned
model based on video thumbnails. In particular,
we use three learning targets: (a) what fraction of 100 evenly-spaced
frames have high responses from a Faster-RCNN hand detector trained on
\cite{Fouhey18}; and (b) what fraction of frames are judged as containing
interaction by human workers;
(c) whether the frames are cartoons. 

These cannot be evaluated on the whole dataset, so we train models
(see supplemental material) that map thumbnails to a prediction of each.
This can be rapidly evaluated at scale on our full dataset's thumbnails
and our final dataset is the intersection between the datasets that are
likely to contain hands and depict interaction, with likely cartoons removed.
These systems are not used in the future and all
subsequent annotations are at a frame-level and are independent 
of video-level filtering mechanism.


\begin{table}[t]
    \centering
    {\footnotesize
    \caption{Comparison of 100DOH with hand datasets. Our dataset is
    far larger and has a rich annotation of contact state with objects.}
    \label{tab:dataimagecompare}
    \begin{tabular}{@{~}l@{~~}c@{~~}c@{~~}c@{~~}c@{~~}c@{~~}c@{~}} \toprule
        Name & \# Im & \# Hands & Side & Contact & Objects & Source \\
        100DOH & 100K & 189.6K & \checkmark & \checkmark & \checkmark & YouTube
        \\
        VLOG \cite{Fouhey18} & 5K & 26.1K & X & Per-Image & X & YouTube
        \\
        VIVA \cite{VIVA} & 5.5K & 13.2K & \checkmark & X & X & Capture
        \\
        Ego \cite{Bambach15} & 4.8K & 15K & \checkmark & X & X & Capture
        \\
        VGG \cite{Mittal11} & 2.7K & 4.2K & X & X & X & Flickr, TV
        \\
        TV-Hand \cite{Narasimhaswamy2019} & 9.5K & 8.6K & X & X & X & TV \\
        COCO-Hand \cite{Narasimhaswamy2019} & 26.5K & 45.7K & X & X & X & Flickr \\
        \bottomrule
    \end{tabular}}
\end{table}

\begin{figure*}[!t]
\centering
\includegraphics[width=0.95\linewidth]{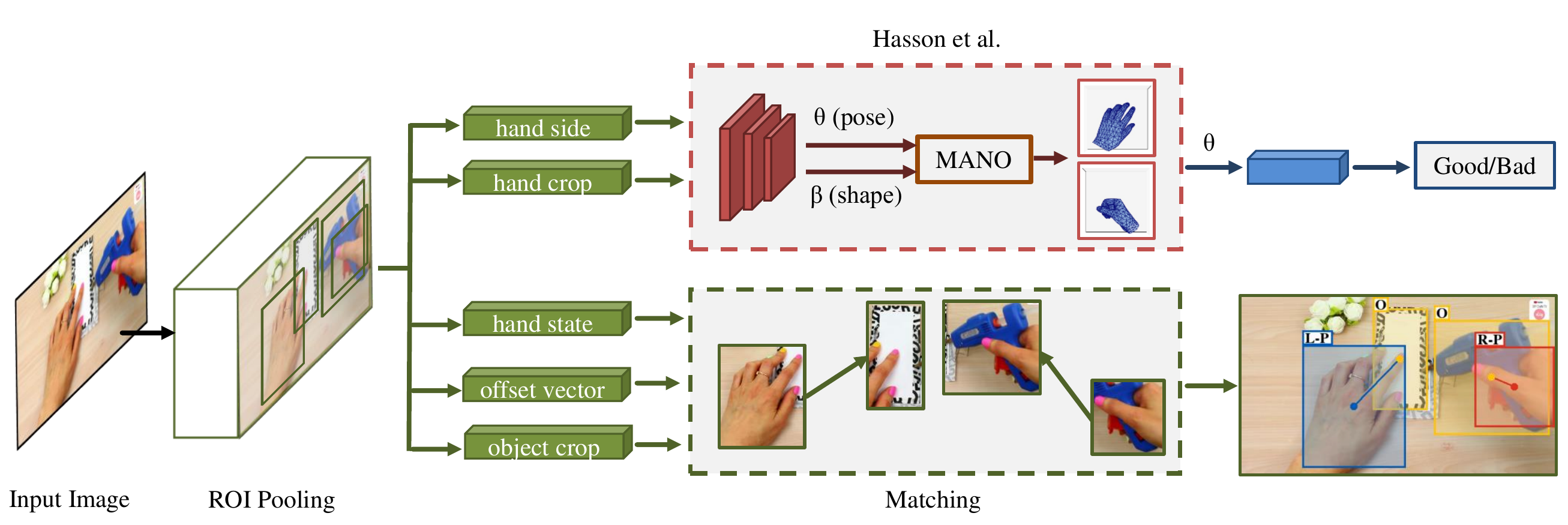}
\caption{Our system can act as a foundation to understand interacting hands on the Internet. 
Our system takes a single RGB image and detects hands (irrespective of scale) and for every hand predicts: a box, side, contact state, and a box around
the object it is touching. We can then (1) obtain a parse of hand state; and (2) use the hand box and side to feed
a reconstruction system like \cite{Hasson19}. To help make better use of Internet reconstructions, we introduce 
a self-supervised system that assesses mesh quality that we train on our data.}
\label{fig:method}
\end{figure*}

\subsection{Image Dataset}

This yields a video dataset -- 100 Days of Hands (100DOH) -- of 27.3K videos across 11 categories
with 131 days of footage of everyday interaction.
We use this to build a new 100K frame-level dataset, 
that is primarily ($\sim85$\%) a subset of 100DOH and ($\sim15$\%) a $3\times$ extended and relabled version
of the hand dataset in \cite{Fouhey18}. We chose randomly among frames,
filtering out (and retaining for later use) images containing no hands. We
include VLOG because we tried building off of VLOG, but realized that we
needed more diverse underlying data.
        

\vspace{2mm}
\noindent {\bf Annotation:}
For every hand in each image, we obtained the following annotations:
{\bf (a)} a bounding box around the hand;
{\bf (b)} side: left / right, which is crucial for mesh reconstruction;
{\bf (c)} the hand contact state (\{no contact, self-contact, other person contact, in contact with portable object,
in contact with a non-portable object\}), which provides insights into what the person is doing;
and {\bf (d)} a bounding box around the object the person is contacting {\it irrespective of name}.
The annotation of non-named bounding boxes is crucial: in-the-wild data is known to have
a heavy tail, dooming categorization.

Annotation began by counting hands, then marking hand bounding boxes and sides simultaneously;
in addition to standard qualification, consensus, and sentinel techniques, hand bounding boxes
were annotated, then verified, then re-annotated if hands were missing. Catching these
missing boxes is crucial since with data of this scale, hand detection performance
can reach a mAP of 90\%. Then hand state and object bounding box were
annotated, again using qualifications, consensus, and sentinels. 
We only included images on which we could get conclusive judgments from workers.
In total, in the 100K images, there are 189.6K hands annotated which are
in contact with 110.1K objects.

\vspace{2mm}
\noindent {\bf Splits:} We split by YouTube uploader id to make a 
80/10/10\% train/val/test split where each uploader appears
in only one split. This split is backwards compatible with the VLOG hand data:
no VLOG test appears in the trainset.

\vspace{2mm}
\noindent {\bf Comparison to existing data:} We compare our dataset in raw
statistics with other comparable datasets of videos in
Table~\ref{tab:datacompare} and image-based datasets for studying hands in
Table~\ref{tab:dataimagecompare} (empirical cross-dataset evaluations
demonstrating the utility of the dataset appear in
Section~\ref{sec:experiments}). While unlabeled, our video dataset 
provides a valuable source of large-scale demonstrations of hands engaged
in interaction. Our image dataset fills a gap of providing
object contact and side information at vast scale. Additionally, 
as seen in the paper, 100DOH hands appear at a wide variety of scales.

\section{Finding Hands \& Objects in Interaction}
\label{sec:method}
Equipped with this data, we show how to build a system that can produce a fine-grained understanding
of the scene.  Our base system (Sec.~\ref{subsec:methodbase}) can predict, from a single image: 
(1) a box around any visible human hands in the scene as well as their {\it side} (left-vs-right) and {\it contact state} (none/self/person/portable/non-portable);
(2) the box of an object the hand is in contact with; and
(3) a {\it link} between each hand and an object it is in contact with. 
Our outputs can be directly plugged into machinery for hand reconstruction (Sec.~\ref{subsec:reconstruction})
\cite{Hasson19,Romero17}, also enabling
(4) a 3D mesh reconstruction; 
(5) and whether that reconstruction was likely correct. 
We believe this output can enable many exciting downstream applications, and we show a proof of concept
of mapping objects to grasps (Sec.~\ref{subsec:grasp}).


\begin{figure*}[!t]
\centering
\includegraphics[width=\linewidth]{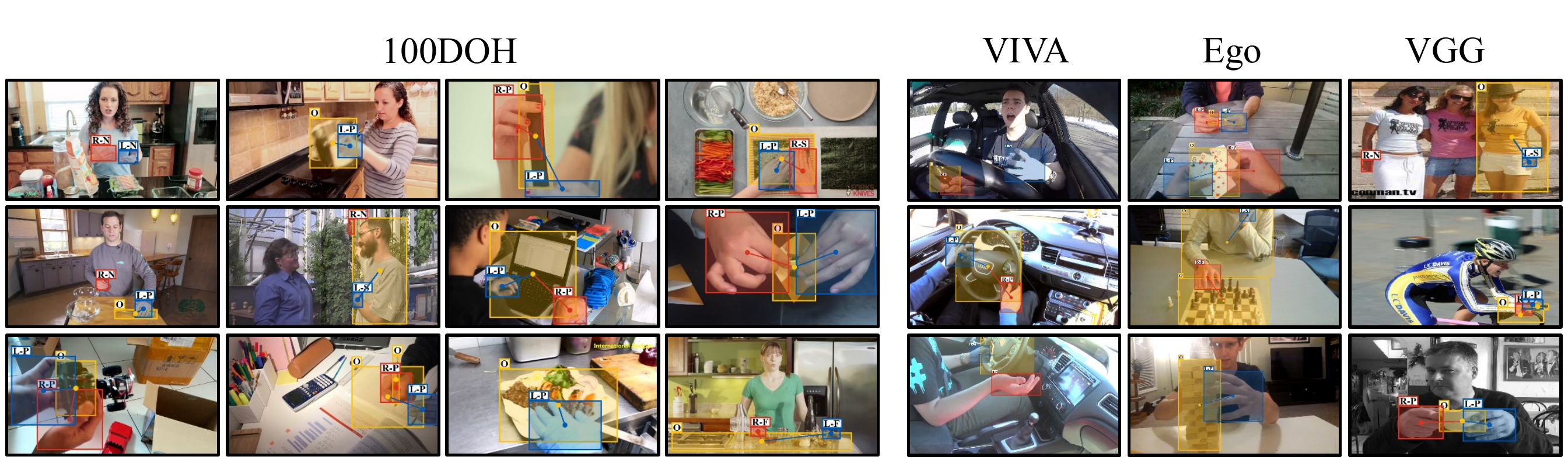}
\caption{Selected results from our full hand state detection system. Here we show our results on 100DOH as well as generalizing
    (untrained) to VIVA \cite{VIVA}, EgoHands \cite{Bambach15}, and VGG \cite{Mittal11}. Our system is
    able to reliably extract hands at a variety of scales, poses, and contexts as well as identify contact
    state and which object is in contact.}
\label{fig:qualitative1}
\end{figure*}

\subsection{Hand and Object Detection}
\label{subsec:methodbase}

We build our system on top of a standard object detection system, Faster-RCNN
\cite{Ren2015} (FRCNN) by adding auxiliary predictions and losses per-bounding box. 
We deliberately chose FRCNN for its reputation as a standard foundation for
detection tasks; we see additional improvements to the base network as orthogonal to our contributions.
Specifically, we build on FRCNN trained to identify two objects -- human hands and contacted
objects. As in standard Faster-RCNN, the network predicts, for each anchor box,
whether the anchor box is an object, what its category is, and bounding box
regression adjustments to the anchor box; these remain unchanged.  We predict a
series of auxiliary outputs directly from the same ROI-pooled features as the
standard classification outputs. We now report these outputs and the losses
we use to train these additional layers:

We predict hand side $\sB \in \mathbb{R}^2$ and contact state $\cB \in \mathbb{R}^5$ 
via two additional fully connected layers. The outputs representing
left-vs-right and \{none / self / other / portable / non-portable\}. Both are trained
by minimizing standard cross-entropy losses $L_\textrm{side}$ and $L_\textrm{state}$.

To link up boxes between hands and objects,
we predict an association from a hand to an object, similar to 
Gkioxari et al. \cite{Gkioxari18}, by predicting an
{\it offset vector}, 
factored into a unit vector
$\vB \in \mathbb{R}^2$ plus a magnitude $m \in \mathbb{R}$ by two fully
connected layers. Given the ground-truth vector between the center of the
bounding box of a hand to the center of the bounding box of the object the hand is
contacting, we write it as a unit vector $\vB' \in \mathbb{R}^2$ and magnitude $m \in \mathbb{R}$.
We minimize the distance between the 
two vectors 
$L_\textrm{ori}(\vB,\vB') = ||\vB - \vB'||^2_2$
as well as the
squared difference between the magnitudes $L_\textrm{mag}(m,m') = (m-m')^2$. Formulating the relationship
as predicting an object per-hand allows multiple hands to contact the same object; while it does preclude
a hand contacting multiple objects, we find this is rarer and leave it to future work.

We obtain a final discrete parse in terms of a set of hands in contact/correspondence with a set of objects
through a greedy optimization on network output. Given a new image, we infer
all the hand and object boxes, as well as their side and contact scores and
association vector. We convert these soft predictions into a discrete
prediction by suppressing unlikely hand/object detections and then
associating each confident hand with the object whose center closest
matches the hand's bounding box center plus its offset vector. 

\vspace{2mm}
\par \noindent {\bf Training details.}
The standard FRCNN losses are minimized as usual; we minimize $L_\textrm{side}$, $L_\textrm{state}$
over anchor boxes corresponding to ground-truth hands and $L_\textrm{ori}$ and $L_\textrm{mag}$ over
anchor boxes corresponding to ground-truth hands in contact.
We scale the loss terms to handle wide variance in the loss scale but otherwise did not tune loss
scales (details in supplemental material). We use a ResNet-101 \cite{He2015} backbone, initialized with Imagenet \cite{ILSVRC15} 
and train it for 8 epochs with a learning rate of $10^{-3}$ with batch size of 1.

\subsection{Applications to Reconstruction}
\label{subsec:reconstruction}

Our system, out-of-the-box, directly enables the large-scale automatic
deployment of techniques for mapping hands to 3D meshes which supplement our
outputs. As a concrete demonstration, we build off of the technique of
\cite{Hasson19} that maps images to the MANO \cite{Romero17} low-dimensional
parameterization of hands via a Resnet-18 \cite{He2015}; this parameterization
comprises $[\thetaB,\betaB]$ representing hand pose and shape, which can be
converted to a 3D mesh via the differentiable MANO model.  Our system provides
the necessary inputs (locations {\it plus side}); building a more complex
system that integrates with the detection system is an interesting future
direction and technically feasible but beyond the scope of a single paper. 


While this enables many interesting downstream tasks, these tasks would
be harmed by incorrect reconstructions and so we present a simple technique 
for recognizing these failures.
We use the ideas of checking a network's
equivariance as a signal for confidence from \cite{Bahat2018}. Specifically,
given an image, we reconstruct the hand from six rotated copies of
the image, reproject joints, rotate them and computed the the mean L2 distance
of corresponding joints. We generate these for 3 frames per training video
(70.9K images), sort by consistency and set the examples
in the top 30\% as positives and the bottom 30\% as negatives.
We train a two layer multilayer perceptron (hidden layer sizes 100, 50) on $[\thetaB]$, minimizing the
binary cross-entropy; this
classifier can be run at inference time to quickly identify poorly
estimated frames. We quantify its effectiveness in Section \ref{sec:experimentsReconstruction}.

\subsection{Proof of Concept: From Object to Grasp}
\label{subsec:grasp}

Once we can identify hands in contact in videos and reconstruct them, we can
generate training data for identifying {\it how} hands might contact an object.
After associating hands to tracks, we search our training set for moments in
time where a hand makes contact with an object. On either side are a timestamp
$t_\textrm{before}$ where a hand is not in contact and a timestamp
$t_\textrm{after}$ where the hand is in contact. At $t_\textrm{after}$, our
system provides side, bounding box for both hand and object, a mesh (via
\cite{Hasson19}), and our self-supervised mesh assessment score. We can use the
object box at $t_\textrm{after}$ to crop the image pre-contact at time
$t_\textrm{before}$. We apply a number of filters, including removing examples
with overlapping hands and scenes where the object appeared to move (detected
by change in appearance).

We can then learn a mapping from an image of an uncontacted object to a hand-in-contact. We use
203.4K training samples to build a system. We fine-tuned 
an Imagenet-pretrained \cite{ILSVRC15} Resnet-18 \cite{He2015} capped by a MLP
to predict, for each mesh, hand pose and side, supervised by standard L2 losses 
along with supervision from hand vertices similar to \cite{Hasson19}. We found that this, like many
regression formulations (e.g., see \cite{zhang2016colorful}), averaged out
between the multiple modes. To prevent averaged hands, we generated a 10-hand codebook
from training samples, represented each hand with the nearest of 10 classes
and predicted these. For simplicity, trained another Resnet-18 to predict
these classes, minimizing a cross-entropy loss. 



\section{Experiments}
\label{sec:experiments}
\begin{table}
    \caption{Cross-dataset performance:
    training Faster-RCNN \cite{Ren2015} on our dataset ensures near-equal performance
    to training and testing on other datasets; other datasets 
    usually generalize poorly. For each train set, we report the minimum ratio (across 
    test sets) between its performance and the best-performer. \dandan{Here, TV represents TV-hand plus COCO-Hand.}}
    
    \label{tab:crossdatasethands}

    \resizebox{\ifdim\width>\columnwidth\columnwidth\else\width\fi}{!}{

    \begin{tabular}{l@{~~~}c@{~~~}c@{~~~}c@{~~~}c@{~~~}c@{~~~}c@{~~~}c} 
        \toprule
        Test $\rightarrow$ & 100 & VLOG & VIVA & Ego & VGG & TV+Co & min \\
        Train $\downarrow$ & DOH & \cite{Fouhey18} & \cite{VIVA} & \cite{Bambach15} & \cite{Mittal11} & \cite{Narasimhaswamy2019} & ratio \\

        \midrule
        100DOH  & \cellcolor{CTX0b0}  \textcolor{CTX0f0}{90.1}  & \cellcolor{CTX0b1}  \textcolor{CTX0f1}{86.4}  & \cellcolor{CTX0b2}  \textcolor{CTX0f2}{86.5}  & \cellcolor{CTX0b3}  \textcolor{CTX0f3}{90.8}  & \cellcolor{CTX0b4}  \textcolor{CTX0f4}{73.9}  & \cellcolor{CTX0b5}  \textcolor{CTX0f5}{65.4}  & \cellcolor{CTX0b6}  \textcolor{CTX0f6}{92.9} \\ 
VLOG  & \cellcolor{CTX1b0}  \textcolor{CTX1f0}{78.6}  & \cellcolor{CTX1b1}  \textcolor{CTX1f1}{77.5}  & \cellcolor{CTX1b2}  \textcolor{CTX1f2}{76.6}  & \cellcolor{CTX1b3}  \textcolor{CTX1f3}{83.2}  & \cellcolor{CTX1b4}  \textcolor{CTX1f4}{64.6}  & \cellcolor{CTX1b5}  \textcolor{CTX1f5}{59.2}  & \cellcolor{CTX1b6}  \textcolor{CTX1f6}{81.1} \\ 
VIVA  & \cellcolor{CTX2b0}  \textcolor{CTX2f0}{23.6}  & \cellcolor{CTX2b1}  \textcolor{CTX2f1}{27.7}  & \cellcolor{CTX2b2}  \textcolor{CTX2f2}{90.8}  & \cellcolor{CTX2b3}  \textcolor{CTX2f3}{56.8}  & \cellcolor{CTX2b4}  \textcolor{CTX2f4}{21.5}  & \cellcolor{CTX2b5}  \textcolor{CTX2f5}{10.1}  & \cellcolor{CTX2b6}  \textcolor{CTX2f6}{14.5} \\ 
Ego  & \cellcolor{CTX3b0}  \textcolor{CTX3f0}{40.7}  & \cellcolor{CTX3b1}  \textcolor{CTX3f1}{32.6}  & \cellcolor{CTX3b2}  \textcolor{CTX3f2}{44.9}  & \cellcolor{CTX3b3}  \textcolor{CTX3f3}{90.7}  & \cellcolor{CTX3b4}  \textcolor{CTX3f4}{17.4}  & \cellcolor{CTX3b5}  \textcolor{CTX3f5}{8.0}  & \cellcolor{CTX3b6}  \textcolor{CTX3f6}{11.5} \\ 
VGG  & \cellcolor{CTX4b0}  \textcolor{CTX4f0}{61.4}  & \cellcolor{CTX4b1}  \textcolor{CTX4f1}{61.7}  & \cellcolor{CTX4b2}  \textcolor{CTX4f2}{78.8}  & \cellcolor{CTX4b3}  \textcolor{CTX4f3}{63.0}  & \cellcolor{CTX4b4}  \textcolor{CTX4f4}{79.6}  & \cellcolor{CTX4b5}  \textcolor{CTX4f5}{66.6}  & \cellcolor{CTX4b6}  \textcolor{CTX4f6}{68.1} \\ 
TV+Co  & \cellcolor{CTX5b0}  \textcolor{CTX5f0}{56.2}  & \cellcolor{CTX5b1}  \textcolor{CTX5f1}{61.5}  & \cellcolor{CTX5b2}  \textcolor{CTX5f2}{74.9}  & \cellcolor{CTX5b3}  \textcolor{CTX5f3}{62.4}  & \cellcolor{CTX5b4}  \textcolor{CTX5f4}{77.4}  & \cellcolor{CTX5b5}  \textcolor{CTX5f5}{69.6}  & \cellcolor{CTX5b6}  \textcolor{CTX5f6}{62.4} \\ 

        \bottomrule
    \end{tabular}}
\end{table}

We conduct a series of experiments that aim to quantify:
(a) how well the dataset allows the hand detection relative
to other hand detection datasets? 
(b) how well our full model for hand-state works?
(c) how much our full hand model assists reconstruction?
(d) how well our model infer interacting hands for isolated objects?
We conduct our experiments on both our newly introduced datasets, as
well as on other hand detection datasets \cite{Fouhey18,VIVA,Mittal11,Bambach15}.

\subsection{Hand Bounding-Box}

We evaluate the merit of our new dataset of hands by evaluating cross-dataset performance for hand detection
with a standard fixed detector, and a comparison with full-body pose estimation.

\begin{table}
\centering
\caption{Average Precision when we vary definitions of correct detection. 
Using 15K samples dramatically degrades performance on any category, and
45K samples produces a large drop on getting all outputs correct.
\note{DF: need updated version. } \dandan{Updated. Need to mention threshold here? (0.0 for the first 4, 0.1 for the last 2)}
}
    \label{tab:datascale}
\begin{tabular}{l@{~~~~}c@{~~~~}c@{~~~~}c@{~~~~}c@{~~~~}c@{~~~~}c} \toprule
      & Hand  & Obj     & H+Side    & H+State       & H+O       & All \\ \midrule
    Full  & \cellcolor{SC0b0} \textcolor{SC0f0}{ 89.6} & \cellcolor{SC0b1} \textcolor{SC0f1}{ 63.9} & \cellcolor{SC0b2} \textcolor{SC0f2}{ 78.9} & \cellcolor{SC0b3} \textcolor{SC0f3}{ 64.0} & \cellcolor{SC0b4} \textcolor{SC0f4}{ 46.9} & \cellcolor{SC0b5} \textcolor{SC0f5}{ 38.5}\\ 
45K  & \cellcolor{SC1b0} \textcolor{SC1f0}{ 88.4} & \cellcolor{SC1b1} \textcolor{SC1f1}{ 61.0} & \cellcolor{SC1b2} \textcolor{SC1f2}{ 77.3} & \cellcolor{SC1b3} \textcolor{SC1f3}{ 62.7} & \cellcolor{SC1b4} \textcolor{SC1f4}{ 39.3} & \cellcolor{SC1b5} \textcolor{SC1f5}{ 31.3}\\ 
15K  & \cellcolor{SC2b0} \textcolor{SC2f0}{ 80.9} & \cellcolor{SC2b1} \textcolor{SC2f1}{ 54.2} & \cellcolor{SC2b2} \textcolor{SC2f2}{ 66.8} & \cellcolor{SC2b3} \textcolor{SC2f3}{ 53.3} & \cellcolor{SC2b4} \textcolor{SC2f4}{ 27.3} & \cellcolor{SC2b5} \textcolor{SC2f5}{ 19.7}\\ 

\bottomrule
\end{tabular}
\end{table}

\vspace{2mm}
\noindent {\bf Cross-dataset hand detection analysis:} We begin by training the same base model -- a standard
F-RCNN \cite{Ren2015} with a Resnet-101 \cite{He2015} backbone -- on a number of datasets
and evaluating same and cross-dataset performance. We use F-RCNN for simplicity and due to its widespread use as
a commodity detection system. We only evaluate on datasets where {\it all} hands in a frame
are annotated with boxes \cite{VIVA,Bambach15,Mittal11,Fouhey18}. We do not compare with 
\cite{GarciaHernando18} since it only annotates {\it one} hand (and note
\cite{Damen18} has no hand boxes). We evaluate hand detection results using an
IoU of 0.5 (i.e., PASCAL \cite{Everingham10}) since unlike objects with clear boundaries e.g., fire hydrants, cars, 
precise boundary between hand and wrist is unclear in the wild.

Table~\ref{tab:crossdatasethands} shows that a model trained on 100DOH
generalizes well across datasets and nearly matches performance obtained training and testing on every other
hand datasets: at worst, it obtains 92.9\% of the mAP of training and testing on
the same data (on VGG). VLOG and VGGHands (both gathered from large-scale diverse data)
generalize reasonably well, but far worse than 100DOH with minimum relative mAP of
81.1\% and 68.1\% respectively; models trained on VIVA and EgoHands perform well on
egocentric datasets but generalize poorly to non-egocentric
\dandan{(VIVA performs slightly well on EGO might because the 3rd-person hands in Ego are similar to the hands in VIVA.)} views (which is unsurprising but
worth quantifying). The annotation format of VGG and TV is a quadrilateral rather than an axis-aligned rectangle, and we preprocess the labels to be axis-aligned to make all annotations consistent; this diminishes results slightly. We show some sample cross-dataset results in Figure~\ref{fig:qualitative1}.

\vspace{2mm}
\noindent {\bf Comparison to full-body pose estimation:} One common question (also asked by
\cite{Narasimhaswamy2019}) is whether we need specialized hand-detection given the success of
full-body pose estimation systems and datasets \cite{Cao2017}. We evaluate this by computing
precision/recall for hand detection using OpenPose \cite{Cao2017}.
We convert body joint configuration and hand detection to a common
evaluation scheme by defining true positives as: (\cite{Cao2017}) when a pose
detector places the hand (estimated as $\wB+0.2(\wB-\eB)$ where $\wB$ and $\eB$ are
wrist and elbow locations) within a ground truth box with
the same center but twice the width and height; ({\it ours}) when the bounding
box detector puts its center point inside the ground-truth box (a higher accuracy standard). 
Due to truncated people, \cite{Cao2017} achieves a low average precision of 
around 43.0\% and effectively maxes out recall at 49.5\%.
At this recall, a Faster RCNN still has a precision of 99.7\%. While current
pose estimators trained on current datasets are effective when the body is mainly visible, 
dedicated hand detectors appear to still be necessary. \dandan{R2: openpose uses diff trainset}

\vspace{2mm}
\noindent {\bf Statistical Baselines:} We additionally test whether the dataset
can be solved with simple statistical baselines. We computed the median box
for all hands as well as a median box for left and right hands. These get an 
AP of 0.08\% and 0.11\%, which shows that the hands are widely distributed
across the image and vary in size.

\subsection{Full Hand State}

We next evaluate our full hand state detection system in
isolation. Here, we show that the scale of our datasets is beneficial by
comparing with results obtained by training on smaller subsets of the data.

\begin{figure}
    \includegraphics[width=\linewidth]{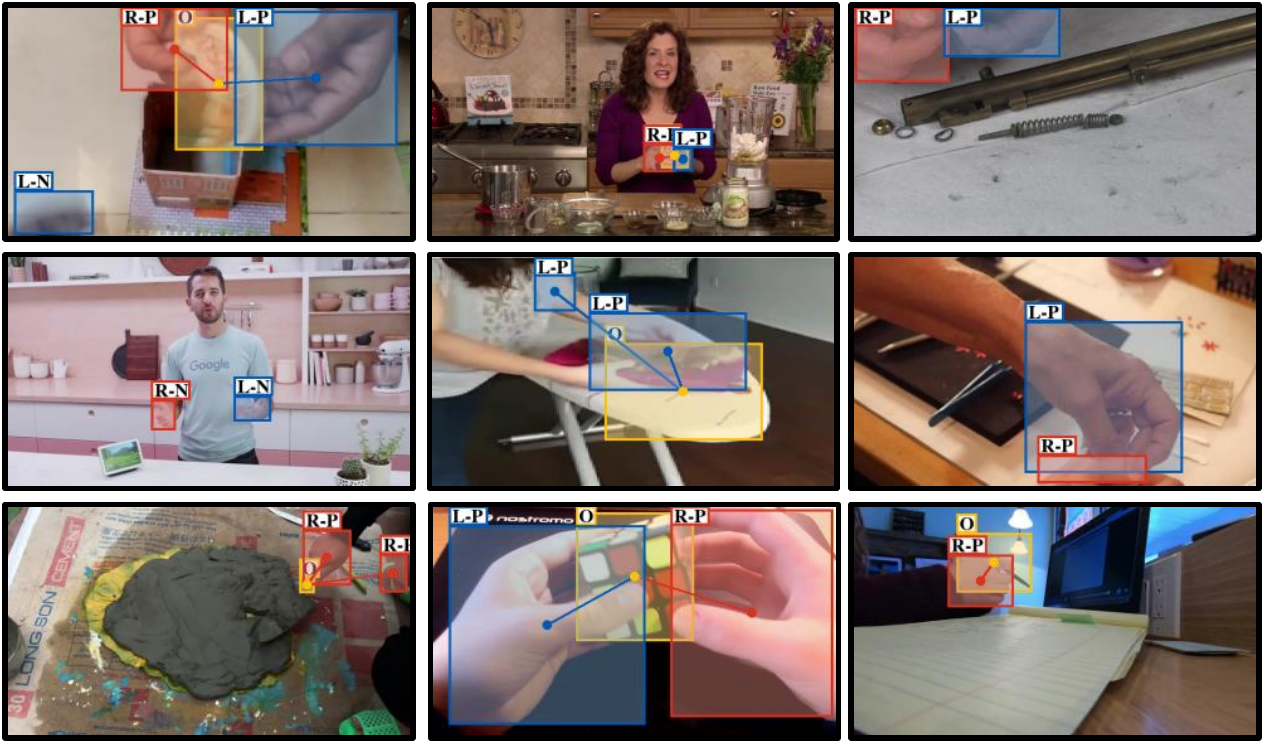}
    \caption{{\bf Random} results on video frames from our 100 DOH dataset. Our approach detects
    hands reliably in a variety of scales, configurations, and contexts.}
    \label{fig:random}
\end{figure}

\vspace{2mm}
\noindent {\bf Qualitative Results:}
We show some results of the full system in Figures \ref{fig:qualitative1} and \ref{fig:random}.
In general, our system does a good job at recognizing hands and sides despite 
a wide variety of scales and contexts present in the data. While it often gets
the full state right, this is a clearly harder task with lots of room for improvement.

\vspace{2mm}
\noindent {\bf Failure modes:} 
Common failure modes include: getting the precise contact state right,
especially when the hand is near an object, which video may improve;
associating the correct object with the hand, especially with multiple people interacting
with multiple objects, which a more complex inference technique might
improve; and getting the correct hand side for egocentric hands (e.g., on \cite{Damen18}), which more
egocentric training data may improve.

\vspace{2mm}
\noindent {\bf Metrics:} We evaluate the full prediction using mAP by modifying
the criterion for a true positive. We begin by evaluating
hand ({\it Hand}) and interacted object ({\it Obj}) individually. 
We then count a hand as a true positive only if it also
has the correct hand side ({\it H+Side}), state ({\it H+State}),
and has the correct object associated with it ({\it H+O}). Finally,
we count a hand as a true positive only 
if it has all correct ({\it All}).

\vspace{2mm}
\noindent {\bf Quantitative Results:} \dandan{Need to update according to Table 4.}
We are the first to tackle this problem, and therefore there are no 
methods to compare with. We therefore test to what extent our
large-scale data is important and compare to the same method
trained on a subsets of data: 
{\it Full} (90K trainval),
{\it 45K} (50\% of trainval); and 
{\it 15K} (17\% of trainval).
We report results in Table~\ref{tab:datascale}. In any category, 
tripling from 15K to 45K produces large gains, while further
doubling to 90K produces more incremental gains. However, when looking
to correctly identify all hand state, these mistakes combine to yield
a steep performance hit (7\% AP) compared to using the full data.
Together, these underscore the need for large-scale data, especially
as one looks to tackle tasks requiring correct estimation of 
{\it multiple} aspects of hands.

\vspace{2mm}
\noindent {\bf Analysis as a Function of Scale:} We evaluate the performance on 
different hand scales. We separate images into different bins according to the average hand size
(measured as the square-root of the percent of pixels) and evaluate each bin.  
Tiny hands are naturally harder to find and hand AP rapidly goes up from 78.2\%
to 90.3\% as scale goes from 10\% to 20\%; performance however remains stable
until 70\%, where it slightly drops. Additional results appear in the supplemental.

\subsection{Hand State for Reconstruction}
\label{sec:experimentsReconstruction}

One of the exciting outcomes of having a system that can reliably
identify hand state is that it directly enables automatically applying
mesh reconstruction techniques to consumer videos. We present two
experiments that assess our method's contributions to this future
-- identifying side and our self-supervised mesh assessment technique. 
We use human judgments to evaluate success.

\vspace{2mm}
\noindent {\bf Data:} We use 2K images from the test set. Our system produced 3,861
detections, which we reconstructed using \cite{Hasson19} for both the correct
hand side and the incorrect hand side resulting in 7,722 meshes. Crowdsourced
workers re-annotated the detected hand to preclude mistakes 
and then assessed each mesh five times as correct/incorrect (definitions in the
supplemental). Workers were deliberately {\bf not} told to inspect
sides of hands. Workers passed a qualification test; we used sentinels to
monitor performance; and results from all reconstructions were annotated
simultaneously and in randomized order.

\begin{figure}
    \includegraphics[width=\linewidth]{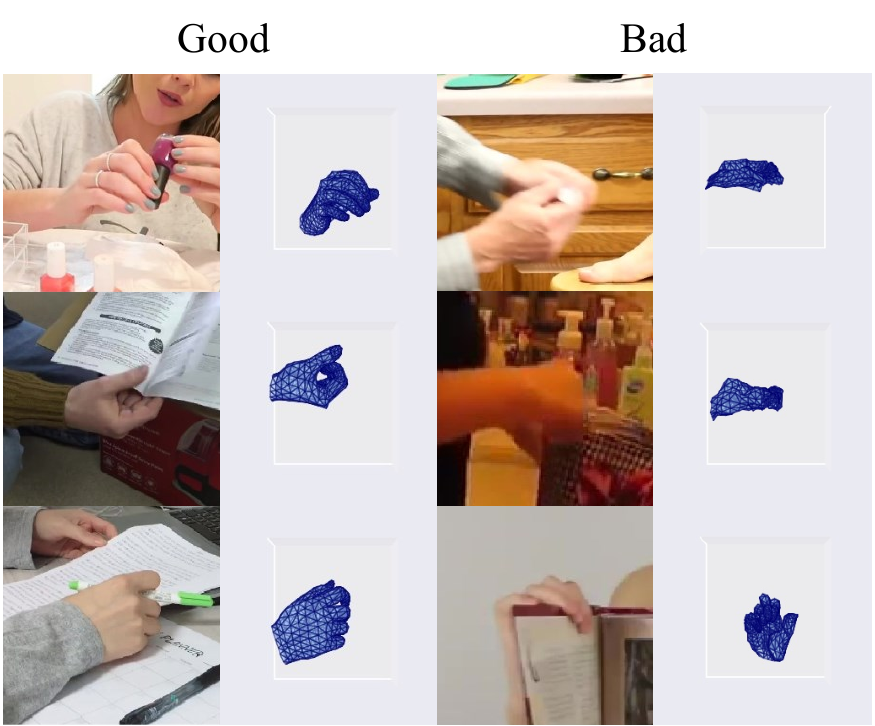}
    \caption{Mesh reconstruction and assessment. Our system detects the location and side 
    of hands in Internet videos, that \cite{Hasson19} uses to predict a mesh. Many 
    predictions do not succeed. Our self-supervised system for mesh assessment
    identifies plausible (useful for downstream tasks) and implausible hands
    (worth discarding). 
    }
    \label{fig:goodbad}
\end{figure}

\vspace{2mm}
\noindent {\bf Quantitative Results (Side):} 
We first tested whether having the side was important -- an alternate hypothesis
is that MANO might repurpose thumbs for pinkies, for instance.  We show a few
select qualitative examples in Figure \ref{fig:goodbad}. Unsurprisingly,
despite not being told to examine side, workers were far more likely to
think hands reconstructed with the {\it detected} side were correct
(57.8\%) compared to the opposite side (29.1\%).  

\vspace{2mm}
\noindent {\bf Quantitative Results (Quality):} We then used this data to evaluate
whether we can successfully identify correct reconstructions. 
We binarized worker judgments
by majority vote and computed AUROC. The proposed method
(a MLP trained on positives/negatives identified by self-consistency)
obtains an AUROC of 90\% on this data. We compared with a two baselines to put
this result in context. Gaussian N\"aive Bayes on the same training data does
similarly (89\%), showing that the positive/negative labels are important, not
the learning algorithm. Simply fitting a multivariate Gaussian on all generated
hands and using the log-likelihood does far worse (60\%), which underscores the
importance of the labels.

%

\subsection{Future prediction}
\label{sec:experimentsFuture}

We took our networks trained on the training set of 100DOH and tested
them on videos from the test set, finding points at which contact changes.
We then reconstruct 3K examples, showing both qualitative results and computing
quantitative results via human judgment.

\begin{figure}[t]
    \includegraphics[width=\linewidth]{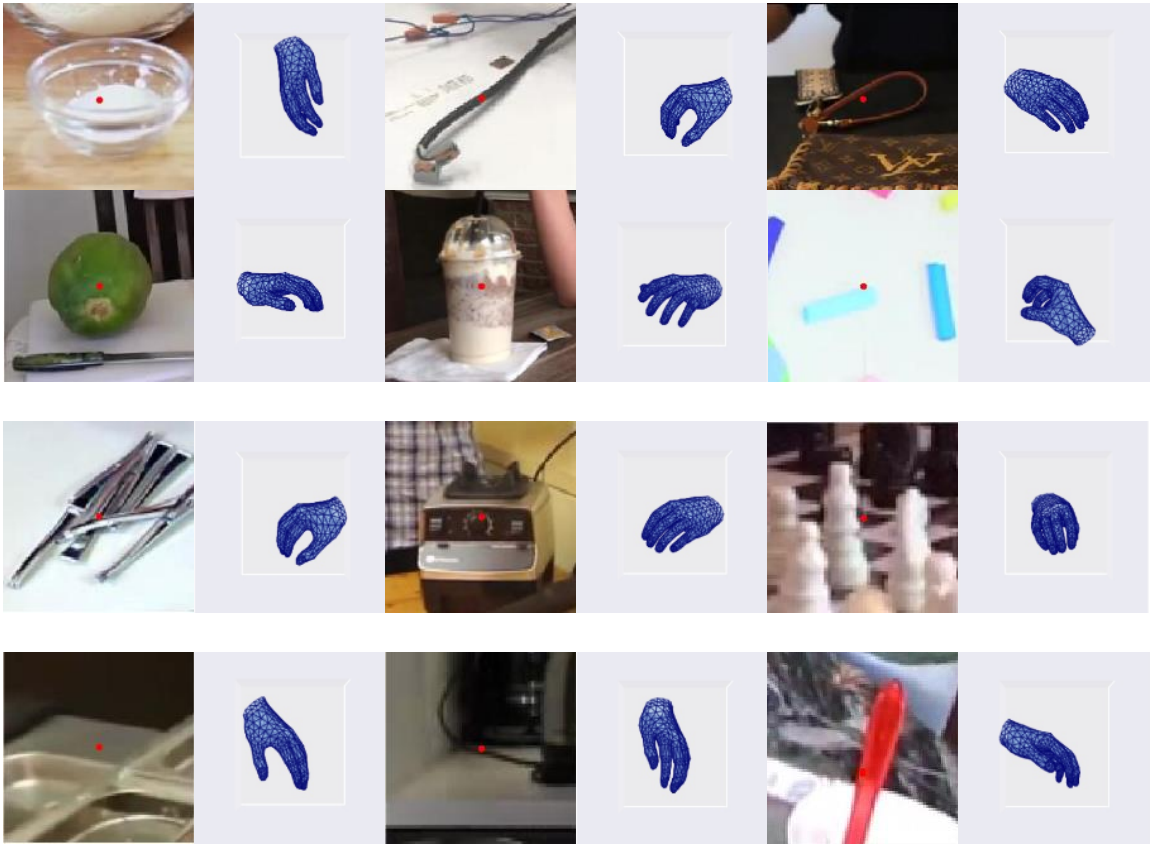}
    \caption{Hand prediction results. Our system enables extracting pairs of images of uncontacted
    objects and {\it good} reconstructed meshes. We show results of a system trained to infer
    meshes from images. (Rows 1,2): Selected Results. (Row 3,4): Random results that crowd workers liked/preferred over a random
    grasp (Row 3) or did not like (preferring random grasp over it).}
    \label{fig:handpred}
\end{figure}

\vspace{2mm}
\noindent {\bf Qualitative Results:}
We show a few select qualitative examples in Figure \ref{fig:handpred}. Overall
we observe that our method often does a good job of identifying
the angle from which the hand should grasp the object. While our approach
often finds plausible grasps, the myriad of ways a human {\it can} grasp
an object and difficulty of predicting a full mesh makes this a challenging task.
\dandan{R1: why only single affordance?}
\note{df: ignoring this}

\vspace{2mm}
\noindent {\bf Human Judgment:}
We showed 3K results to crowd-workers in a two-choice test, comparing the
result to a random hand from the training set to examine if our
system extracts the signal. Note that random is very frequently correct by chance since
usually very many grasps suffice (consider how many ways your hand can touch a soda can).
Workers selected which they thought was more plausible given the image.
Presentation order was randomized, and we employed qualifications and 
sentinels; examples where workers could not come to an agreement were considered
ties. Of the 60\% with a conclusive result (some inconclusive results are
due to the input not depicting a clear object), our system was preferred 72\% of the time.

\section{Conclusion}
We have presented a method for obtaining information about hand contact state
in the scene, a large-scale dataset for training this method, and demonstrated
applications of our technique. We are barely scratching the surface in terms
of what can be learned in the world of large-scale Internet video and we
hope that our rich representation can help the field collectively 
explore this area.

\vspace{2mm}
\noindent {\bf Acknowledgments}: This work was supported by: the Advanced Machine Learning
Collaborative Grant from Procter \& Gamble in collaboration with Matthew Barker, PhD;
and a gift from Nokia Solutions and Networks Oy.

{\small
\bibliographystyle{ieee_fullname}
\bibliography{local}

\begin{thebibliography}{10}\itemsep=-1pt

\bibitem{VIVA}
The vision for intelligent vehicles and applications {(VIVA)} challenge,
  laboratory for intelligent and safe automobiles, {UCSD}.
\newblock http://cvrr.ucsd.edu/vivachallenge/.

\bibitem{Akizuki2018}
Shuichi Akizuki and Yoshimitsu Aoki.
\newblock Tactile logging for understanding plausible tool use based on human
  demonstration.
\newblock In {\em BMVC}, 2018.

\bibitem{Alayrac2016}
Jean-Baptiste Alayrac, Piotr Bojanowski, Nishant Agrawal, Ivan Laptev, Josef
  Sivic, and Simon Lacoste-Julien.
\newblock Unsupervised learning from narrated instruction videos.
\newblock In {\em CVPR}, 2016.

\bibitem{Bahat2018}
Yuval Bahat and Gregory Shakhnarovich.
\newblock Confidence from invariance to image transformations.
\newblock {\em CoRR}, abs/1804.00657, 2018.

\bibitem{Bambach15}
Sven Bambach, Stefan Lee, David Crandall, and Chen Yu.
\newblock Lending a hand: Detecting hands and recognizing activities in complex
  egocentric interactions.
\newblock In {\em ICCV}, 2015.

\bibitem{Baradel18}
Fabien Baradel, Natalia Neverova, Christian Wolf, Julien Mille, and Greg Mori.
\newblock Object level visual reasoning in videos.
\newblock In {\em ECCV}, 2018.

\bibitem{Brahmbhatt19}
Samarth Brahmbhatt, Cusuh Ham, Charlie Kemp, and James Hays.
\newblock Contactdb: Analyzing and predicting grasp contact via thermal
  imaging.
\newblock In {\em CVPR}, 2019.

\bibitem{Cao2017}
Zhe Cao, Tomas Simon, Shih-En Wei, and Yaser Sheikh.
\newblock Realtime multi-person 2d pose estimation using part affinity fields.
\newblock In {\em CVPR}, 2017.

\bibitem{Chao2018}
Yu-Wei Chao, Yunfan Liu, Xieyang Liu, Huayi Zeng, and Jia Deng.
\newblock Learning to detect human-object interactions.
\newblock In {\em Proceedings of the IEEE Winter Conference on Applications of
  Computer Vision}, 2018.

\bibitem{Chao15}
Yu-Wei Chao, Zhan Wang, Yugeng He, Jiaxuan Wang, and Jia Deng.
\newblock Hico: A benchmark for recognizing human-object interactions in
  images.
\newblock In {\em ICCV}, 2015.

\bibitem{Damen18}
Dima Damen, Hazel Doughty, Giovanni~Maria Farinella, Sanja Fidler, Antonino
  Furnari, Evangelos Kazakos, Davide Moltisanti, Jonathan Munro, Toby Perrett,
  Will Price, and Michael Wray.
\newblock Scaling egocentric vision: The epic-kitchens dataset.
\newblock In {\em ECCV}, 2018.

\bibitem{Everingham10}
M. Everingham, L. Van~Gool, C.~K.~I. Williams, J. Winn, and A. Zisserman.
\newblock The {PASCAL Visual Object Classes (VOC)} challenge.
\newblock {\em IJCV}, 88(2):303--338, 2010.

\bibitem{Fang2018}
Kuan Fang, Te-Lin Wu, Daniel Yang, Silvio Savarese, and Joseph~J. Lim.
\newblock Demo2vec: Reasoning object affordances from online videos.
\newblock In {\em CVPR}, 2018.

\bibitem{Fouhey18}
David~F. Fouhey, Weicheng Kuo, Alexei~A. Efros, and Jitendra Malik.
\newblock From lifestyle {VLOGs} to everyday interactions.
\newblock In {\em CVPR}, 2018.

\bibitem{GarciaHernando18}
Guillermo Garcia-Hernando, Shanxin Yuan, Seungryul Baek, and Tae-Kyun Kim.
\newblock First-person hand action benchmark with rgb-d videos and 3d hand pose
  annotations.
\newblock In {\em CVPR}, 2018.

\bibitem{Gibson79}
J. Gibson.
\newblock {\em The ecological approach to visual perception}.
\newblock Boston: Houghton Mifflin, 1979.

\bibitem{Gkioxari18}
Georgia Gkioxari, Ross Girshick, Piotr Dollar, and Kaiming He.
\newblock Detecting and recognizing human-object interactions.
\newblock In {\em CVPR}, 2018.

\bibitem{Gu2018}
Chunhui Gu, Chen Sun, Sudheendra Vijayanarasimhan, Caroline Pantofaru, David~A.
  Ross, George Toderici, Yeqing Li, Susanna Ricco, Rahul Sukthankar, Cordelia
  Schmid, and Jitendra Malik.
\newblock {AVA:} {A} video dataset of spatio-temporally localized atomic visual
  actions.
\newblock In {\em CVPR}, 2018.

\bibitem{Gupta15}
Saurabh Gupta and Jitendra Malik.
\newblock Visual semantic role labeling.
\newblock {\em arXiv preprint arXiv:1505.04474}, 2015.

\bibitem{Hamer2010}
Henning Hamer, Juergen Gall, Thibaut Weise, and Luc~Van Gool.
\newblock An object-dependent hand pose prior from sparse training data.
\newblock In {\em CVPR}, 2010.

\bibitem{Hasson19}
Yana Hasson, G{\"u}l Varol, Dimitrios Tzionas, Igor Kalevatykh, Michael~J.
  Black, Ivan Laptev, and Cordelia Schmid.
\newblock Learning joint reconstruction of hands and manipulated objects.
\newblock In {\em CVPR}, 2019.

\bibitem{He2015}
Kaiming He, Xiangyu Zhang, Shaoqing Ren, and Jian Sun.
\newblock Deep residual learning for image recognition.
\newblock In {\em CVPR}, 2016.

\bibitem{Kay17}
Will Kay, Joao Carreira, Karen Simonyan, Brian Zhang, Chloe Hillier, Sudheendra
  Vijayanarasimhan, Fabio Viola, Tim Green, Trevor Back, Paul Natsev, Mustafa
  Suleyman, and Andrew Zisserman.
\newblock The kinetics human action video dataset.
\newblock {\em CoRR}, abs/1705.06950, 2017.

\bibitem{miech2019howto100m}
Antoine Miech, Dimitri Zhukov, Jean-Baptiste Alayrac, Makarand Tapaswi, Ivan
  Laptev, and Josef Sivic.
\newblock Howto100m: Learning a text-video embedding by watching hundred
  million narrated video clips.
\newblock {\em arXiv preprint arXiv:1906.03327}, 2019.

\bibitem{Mittal11}
A. Mittal, A. Zisserman, and P.~H.~S. Torr.
\newblock Hand detection using multiple proposals.
\newblock In {\em BMVC}, 2011.

\bibitem{MonfortMoments2019}
Mathew Monfort, Alex Andonian, Bolei Zhou, Kandan Ramakrishnan, Sarah~Adel
  Bargal, Tom Yan, Lisa Brown, Quanfu Fan, Dan Gutfruend, Carl Vondrick, et~al.
\newblock Moments in time dataset: one million videos for event understanding.
\newblock {\em IEEE Transactions on Pattern Analysis and Machine Intelligence},
  pages 1--8, 2019.

\bibitem{Nagarajan2019}
Tushar Nagarajan, Christoph Feichtenhofer, and Kristen Grauman.
\newblock Grounded human-object interaction hotspots from video.
\newblock In {\em ICCV}, 2019.

\bibitem{Narasimhaswamy2019}
Supreeth Narasimhaswamy, Zhengwei Wei, Yang Wang, Justin Zhang, and Minh Hoai.
\newblock Contextual attention for hand detection in the wild.
\newblock In {\em ICCV}, 2019.

\bibitem{Ren2015}
Shaoqing Ren, Kaiming He, Ross Girshick, and Jian Sun.
\newblock Faster {R-CNN}: Towards real-time object detection with region
  proposal networks.
\newblock In {\em {NIPS}}, 2015.

\bibitem{Rogez15}
G. Rogez, J. Supancic, and D. Ramanan.
\newblock Understanding everyday hands in action from rgb-d images.
\newblock In {\em ICCV}, 2015.

\bibitem{Romero17}
J. Romero, D. Tzionas, and M.~J. Black.
\newblock Embodied hands: Modeling and capturing hands and bodies together.
\newblock {\em {ACM} Transactions on Graphics, {(Proc. SIGGRAPH Asia)}}, 36(6),
  2017.

\bibitem{ILSVRC15}
Olga Russakovsky, Jia Deng, Hao Su, Jonathan Krause, Sanjeev Satheesh, Sean Ma,
  Zhiheng Huang, Andrej Karpathy, Aditya Khosla, Michael Bernstein,
  Alexander~C. Berg, and Li Fei-Fei.
\newblock {ImageNet Large Scale Visual Recognition Challenge}.
\newblock {\em IJCV}, pages 1--42, April 2015.

\bibitem{Sigurdsson2016}
Gunnar~A. Sigurdsson, G{\"u}l Varol, Xiaolong Wang, Ali Farhadi, Ivan Laptev,
  and Abhinav Gupta.
\newblock Hollywood in homes: Crowdsourcing data collection for activity
  understanding.
\newblock In {\em ECCV}, 2016.

\bibitem{Spurr2018}
A. Spurr, J. Song, S. Park, and O. Hilliges.
\newblock Cross-modal deep variational hand pose estimation.
\newblock In {\em CVPR}, 2018.

\bibitem{Sridhar2016}
Srinath Sridhar, Franziska Mueller, Michael Zollh\"ofer, Dan Casas, Antti
  Oulasvirta, and Christian Theobalt.
\newblock Real-time joint tracking of a hand manipulating an object from rgb-d
  input.
\newblock In {\em ECCV}, 2016.

\bibitem{Tekin19}
Bugra Tekin, Federica Bobo, and Marc Pollefeys.
\newblock H+o: Unified egocentric recognition of 3d hand-object poses and
  interactions.
\newblock In {\em CVPR}, 2019.

\bibitem{Torralba2011}
A. Torralba and A.~A. Efros.
\newblock Unbiased look at dataset bias.
\newblock In {\em CVPR}, 2011.

\bibitem{zhang2016colorful}
Richard Zhang, Phillip Isola, and Alexei~A Efros.
\newblock Colorful image colorization.
\newblock In {\em ECCV}, 2016.

\bibitem{Zhou2018}
Luowei Zhou, Chenliang Xu, and Jason~J. Corso.
\newblock Towards automatic learning of procedures from web instructional
  videos.
\newblock In {\em AAAI}, 2018.

\bibitem{Zhukov19}
Dimitri Zhukov, Jean-Baptiste Alayrac, Ramazan~Gokberk Cinbis, David Fouhey,
  Ivan Laptev, and Josef Sivic.
\newblock Cross-task weakly supervised learning from instructional videos.
\newblock In {\em CVPR}, 2019.

\bibitem{Zimmerman2017}
C. Zimmerman and T. Brox.
\newblock Learning to estimate 3d hand pose from single rgb images.
\newblock In {\em ICCV}, 2017.

\bibitem{Freihand2019}
Christian Zimmermann, Duygu Ceylan, Jimei Yang, Bryan Russell, Max Argus, and
  Thomas Brox.
\newblock Freihand: A dataset for markerless capture of hand pose and shape
  from single rgb images.
\newblock In {\em IEEE International Conference on Computer Vision (ICCV)},
  2019.

\end{thebibliography}
}

\end{document}